\documentclass[10pt,twocolumn,letterpaper]{article}

\usepackage{cvpr}              

\definecolor{cvprblue}{rgb}{0.21,0.49,0.74}
\usepackage[pagebackref,breaklinks,colorlinks,citecolor=cvprblue]{hyperref}
\usepackage{graphicx}
\usepackage{amsmath}
\usepackage{amssymb}
\usepackage{booktabs}
\usepackage{diagbox}
\usepackage{microtype}      
\usepackage{graphicx}
\usepackage{float}
\usepackage{bbding}
\usepackage{enumitem}
\def\confName{CVPR}
\def\confYear{2025}

\usepackage{epsfig}
\usepackage{multicol}
\usepackage{multirow}

\usepackage[capitalize]{cleveref}
\crefname{section}{Sec.}{Secs.}
\Crefname{section}{Section}{Sections}
\Crefname{table}{Table}{Tables}
\crefname{table}{Tab.}{Tabs.}

\def\authorBlock{
    Guanglu Dong \qquad
    Xiangyu Liao \qquad
    Mingyang Li   \qquad
    Guihuan Guo \qquad
    Chao Ren\thanks{Corresponding author} 
     \\

    College of Electronics and Information Engineering, Sichuan University, China  \\

	{\tt\small \{dongguanglu, liaoxiangyu1, 2024222055110, guoguihuan\}@stu.scu.edu.cn, } 
	{\tt\small chaoren@scu.edu.cn}
    
}

\title{Exploring Semantic Feature Discrimination for Perceptual Image Super-Resolution and Opinion-Unaware No-Reference Image Quality Assessment}
\author{\authorBlock}

\begin{document}
\maketitle
\renewcommand{\thefootnote}{\fnsymbol{footnote}}

\begin{abstract}
Generative Adversarial Networks (GANs) have been widely applied to image super-resolution (SR) to enhance the perceptual quality. However, most existing GAN-based SR methods typically perform coarse-grained discrimination directly on images and ignore the semantic information of images, making it challenging for the super resolution networks (SRN) to learn fine-grained and semantic-related texture details. To alleviate this issue, we propose a semantic feature discrimination method, SFD, for perceptual SR. Specifically, we first design a feature discriminator (Feat-D), to discriminate the pixel-wise middle semantic features from CLIP, aligning the feature distributions of SR images with that of high-quality images. Additionally, we propose a text-guided discrimination method (TG-D) by introducing learnable prompt pairs (LPP) in an adversarial manner to perform discrimination on the more abstract output feature of CLIP, further enhancing the discriminative ability of our method. With both Feat-D and TG-D, our SFD can effectively distinguish between the semantic feature distributions of low-quality and high-quality images, encouraging SRN to generate more realistic and semantic-relevant textures. Furthermore, based on the trained Feat-D and LPP, we propose a novel opinion-unaware no-reference image quality assessment (OU NR-IQA) method, SFD-IQA, greatly improving OU NR-IQA performance without any additional targeted training. Extensive experiments on classical SISR, real-world SISR, and OU NR-IQA tasks demonstrate the effectiveness of our proposed methods. Code is available at \href{https://github.com/GuangluDong0728/SFD}{https://github.com/GuangluDong0728/SFD}.
\end{abstract}

\section{Introduction}
\label{sec:intro}

\begin{figure}[t]
  \centering
  \setlength{\abovecaptionskip}{0.15cm}
   \includegraphics[width=1.0\linewidth]{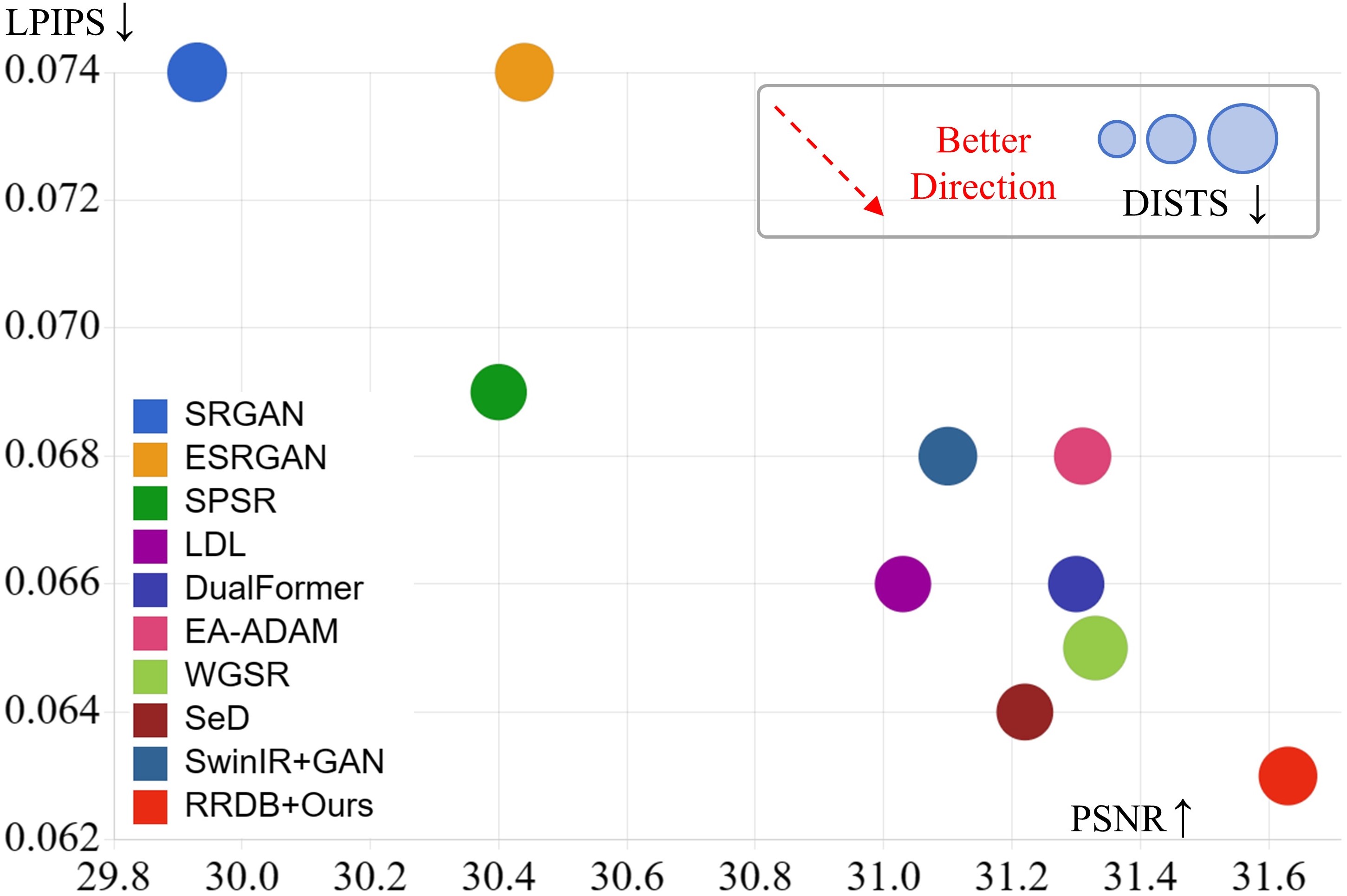}

   \caption{Perception-distortion trade-off comparison between our proposed SFD and other SOTA GAN-based SISR methods.}
   \label{fig.first}
\end{figure}

\begin{figure}[t]
  \centering
  \setlength{\abovecaptionskip}{0.1cm}
   \includegraphics[width=1.0\linewidth]{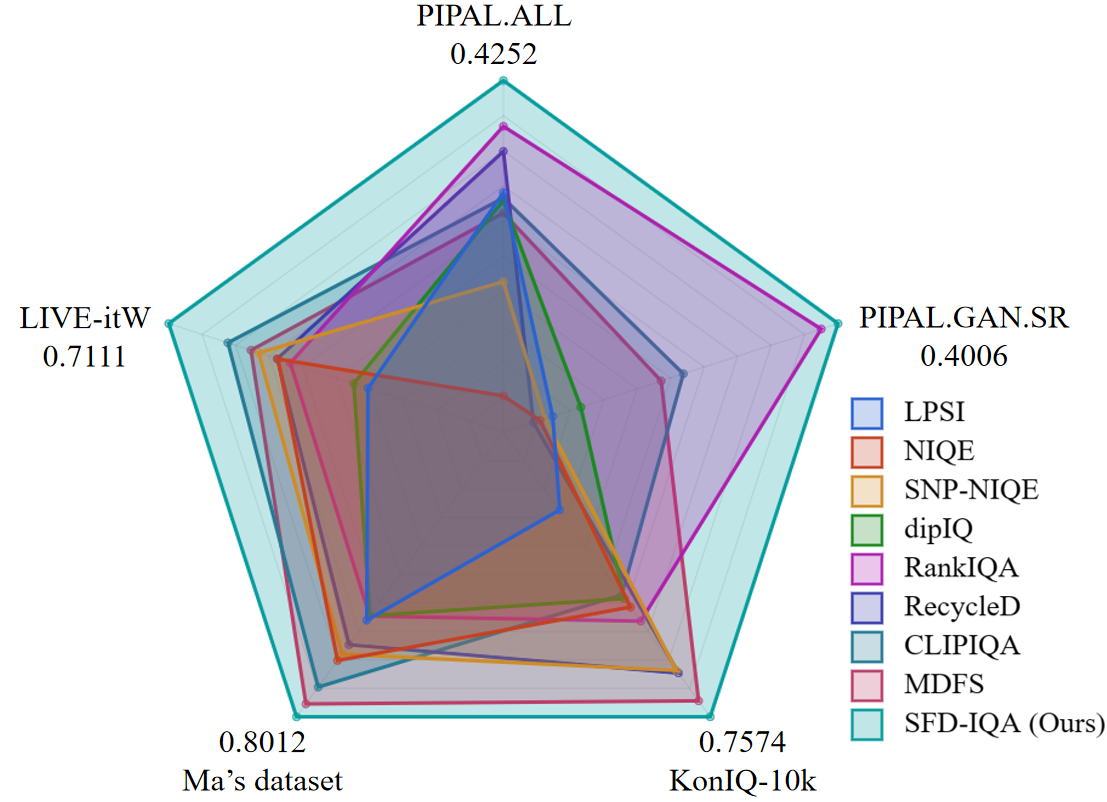}

   \caption{Relative PLCC comparation between the proposed SFD-IQA and other OU NR-IQA methods.}
   \label{fig.second}
\end{figure}

\begin{figure*}[t]
  \centering
  \setlength{\abovecaptionskip}{0.1cm}
   \includegraphics[width=1.0\linewidth]{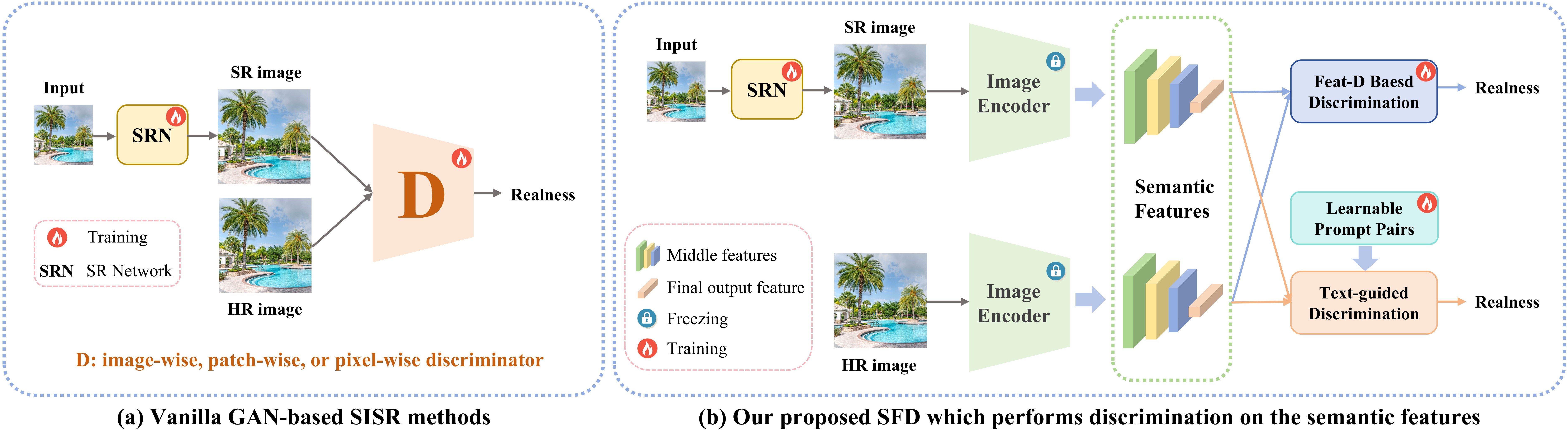}

   \caption{(a) Vanilla GAN-based SISR methods: performing discrimination on images; (b) Our proposed SFD: performing discrimination on the semantic features. Vanilla GAN-based SISR typically uses an image-wise, patch-wise, or pixel-wise discriminator to determine if the distribution of SR images align with that of high-quality real-world images, ignoring the semantic information of images. Such coarse-grained discrimination makes it challenging for the SR network to reconstruct fine-grained, semantic-relevant textures. In contrast, we propose to perform discrimination on the semantic features, encouraging the SR network to generate more realistic semantic textures.}
   \label{fig.differences}
\end{figure*}

As one of the most important tasks in low-level computer vision \cite{JIAYU, zunei_denoising, zunei_SR, promptrestorer, darkchannel, DerainNet, promptir, uformer}, single image super-resolution (SISR) \cite{SRCNN, StableSR, MemNet, ELAN, DBPN, EDSR, DRRN, LapSRN} aims to accurately reconstruct high-resolution (HR) images from the given low-resolution (LR) counterparts. In recent years, with advancements in deep learning, SISR has achieved rapid and substantial progress. Since Dong et al. \cite{SRCNN} first introduce deep learning to SISR task, numerous studies have emerged to explore more advanced network architectures, including CNN-based models \cite{RDN, RCAN, MemNet, EDSR, DRRN, LapSRN} and Transformer-based models \cite{ELAN, SWINIR, transformerSR1, transformerSR2, transformerSR3, transformerSR4}, achieving continuous improvements in fidelity-oriented performance. However, these works primarily employ pixel-wise L1 or MSE losses for the training process, which often produce overly smooth results and fail to generate realistic texture details \cite{SRGAN, PDTradeOff}.

To overcome the above limitations, a series of generative adversarial networks (GANs) ~\cite{GAN} based SISR models~\cite{Real-esrgan, LDL, Sed, UGAN, FFTGAN, SFTGAN_OST, SROOE, SSL, CAL-GAN, SRGAN} have been proposed to generate more realistic images that align more closely with human visual perception by effectively balancing perceptual quality and fidelity. As shown in Figure \ref{fig.differences} (a), most existing GAN-based methods perform discrimination on the images using an image-wise \cite{VGG}, patch-wise \cite{PatchGAN}, or pixel-wise \cite{PixelGAN} discriminator to distinguish between the distributions of SR images and HR images. However, directly discriminating on the images is coarse-grained and lacks semantic awareness. This leads to that the textures of generated images do not match its semantic and often come with artificial artifacts, making it challenging to produce realistic semantic textures. 

In this paper, different from existing works, we propose to perform discrimination on the semantic features, aiming to align the semantic feature distributions of SR images more closely with that of HR images. Considering that CLIP \cite{CLIP} can extract more interpretable, semantic-aware and quality-relevant features \cite{CLIPIQA, Sed, CLIPDenoising, Clipscore}, we select CLIP's image encoder as the semantic feature extractor and propose two methods to perform joint discrimination on the semantic features. Specifically, as shown in Figure \ref{fig.differences} (b), as for the pixel-wise middle semantic features, we first design a feature discriminator, Feat-D, to distinguish between the feature distributions of SR images and HR images, encouraging the semantic feature distributions of SR images to match that of HR images. As for the final output feature which is more abstract and global, we propose a text-guided discrimination method, TG-D, by introducing learnable prompt pairs (LPP) into an adversarial training process to make them sensitive to global image quality and thus able to discriminate the abstract semantic features, further enhancing our method’s overall discriminative capability. With both Feat-D and TG-D, the SR network is able to learn more fine-grained semantic feature distributions, ultimately producing more realistic and semantic-related texture details.

Additionally, existing researches \cite{RecycleD, DualFormer, UGAN} have shown that during the training process where the discriminators in GAN-based SISR learn to distinguish the realness of image distributions, they simultaneously and implicitly learn to assess image quality, making them highly applicable to opinion-unaware no-reference image quality assessment (OU NR-IQA). Simultaneously, recent studies \cite{Clipscore, cliqa} have also shown the advantage of CLIP for OU NR-IQA. In this paper, to further improve OU NR-IQA performance, we propose a novel OU NR-IQA method, SFD-IQA, by integrating the well trained Feat-D and TG-D which conveniently combines both of the above advantages. Thanks to the superior semantic feature discrimination capability of Feat-D and TG-D and the inherent image quality correlation of CLIP, we greatly improve the OU NR-IQA performance without any additional task-specific training on IQA datasets.

As shown in \cref{fig.first}, our SFD demonstrates a better perception-distortion (PD) trade-off on the PSNR-LPIPS-DISTS plane. Compared to other state-of-the-art (SOTA) GAN-based SISR methods, our LPIPS and PSNR scores are the best among all methods with competitive DISTS scores. Additionally, \cref{fig.second} illustrates a comparison between our SFD-IQA and other SOTA OU NR-IQA methods, where SFD-IQA outperforms other methods across both SR IQA datasets and authentically distorted IQA datasets. The main contributions of this work are summarized below:
\begin{itemize}
\item
We propose a semantic feature discrimination (SFD) method for perceptual SISR, which consists of a feature discriminator (Feat-D) based discrimination and a text-guided discrimination (TG-D). Our SFD can effectively distinguish between the CLIP semantic feature distributions of SR and HR images, encouraging SR networks to learn more fine-grained and semantic-related textures, and achieving better perception-distortion (PD) trade-off.
\item
We propose a novel OU NR-IQA method, SFD-IQA, by effectively integrating the two well trained feature discrimination methods. Without any additional task-specific training on IQA datasets, SFD-IQA can significantly enhance the OU NR-IQA performance for both SR images and other authentically distorted images.
\item
Extensive experiments on classical SISR, real-world SISR, and OU NR-IQA tasks demonstrate the effectiveness of our method. SFD can help SR networks achieve better PD trade-off and generate more realistic semantic textures, and SFD-IQA is able to greatly improve OU NR-IQA performance. Moreover, our approach can be easily integrated into other GAN-based image restoration methods.
\end{itemize}

\section{Related Works}
\label{sec:Related Works}
\subsection{Fidelity-Oriented SISR}

Recently, deep learning based single image super-resolution (SISR) ~\cite{SRCNN, DBPN, LapSRN, RCAN, RDN, SeeSR, StableSR, SRGAN} has made tremendous advancements. Since SRCNN~\cite{SRCNN} first introduces CNN into SISR, numerous SISR backbones \cite{RDN, DRRN, MemNet, RCAN, EDSR, DBPN, LapSRN} have continued to emerge to enhance the performance of fidelity-oriented SISR. Specifically, RCAN \cite{RCAN} introduces channel attention into SISR network, and RDN \cite{RDN} leverages dense connection to fully utilize hierarchical features for SISR. Furthermore, based on the success of the vision transformer \cite{ViT} framework, \cite{SWINIR, ELAN, transformerSR1, transformerSR2} further improve the SISR performance. However, in order to maximize fidelity, most of the aforementioned methods only employ pixel-wise loss functions to optimize the training process and ignore the perceptual quality, which often leads to overly smooth results and fails to achieve visually appealing results.

\subsection{Perception-Oriented GAN-based SISR}
To improve perceptual quality, GAN-based SISR~\cite{Sed, WGSR, ESRGAN, LDL, Real-esrgan, DualFormer, BSRGAN, RankSRGAN, FFTGAN, CAL-GAN} has received constant attention in recent years. Specifically, SRGAN \cite{SRGAN} introduces GANs into SR tasks to obtain SR images more in line with human perception. DualFormer \cite{DualFormer} applies an improved spectral discriminator to improve perceptual quality. Sed~\cite{Sed} incorporates the semantic features into the vanilla discriminator to finely distinguish the image distributions. EA-ADAM \cite{EA-Adam} develops a new optimizer from a multi-objective optimization perspective to achieve a better perception-distortion tradeoff. However, due to the lack of semantic-relevant perception, most GAN-based SR networks are difficult to produce fine-grained semantic textures, result in exhibiting artificial artifacts. In this paper, we propose to discriminate the semantic features of CLIP to guide the SR network to learn more fine-grained semantic feature distributions and generate more realistic semantic textures. 

\subsection{Opinion-Unaware NR-IQA}
OU NR-IQA~\cite{QAC, CLIPIQA, NIQE, ContentSep, MDFS, LPSI, RecycleD, dipIQ} aims to evaluate the quality of images without human-labeled data. Traditional OU NR-IQA methods~\cite{NIQE, IL-NIQE, LPSI} are typically based on natural scene statistics (NSS), which estimate distribution parameters to assess quality. Researches on deep learning-based OU NR-IQA are relatively sparse, with most methods centered around learning-to-rank~\cite{Rankiqa,dipIQ} and contrastive learning~\cite{cliqa}. Recently, with the emergence of the pre-trained vision-language model CLIP~\cite{CLIP}, \cite{CLIPIQA, Clipscore} leverage the rich visual-linguistic priors encapsulated in CLIP to evaluate image perceptual quality. Additionally, RecycleD~\cite{RecycleD} proposes to recycle the trained discriminator for GAN-based SISR to predict perceptual quality. Following~\cite{RecycleD}, \cite{DualFormer} and \cite{UGAN} further enhance the IQA ability of the discriminators in GAN-based SISR. However, these discriminator-based methods only rely on the trained semantic-unaware discriminator and ignore the assessment for semantic, which is difficult to achieve satisfactory performance. 

\section{Method and Analyses}
\begin{figure*}[ht]
  \centering
  \setlength{\abovecaptionskip}{0.1cm}
   \includegraphics[width=1.0\linewidth]{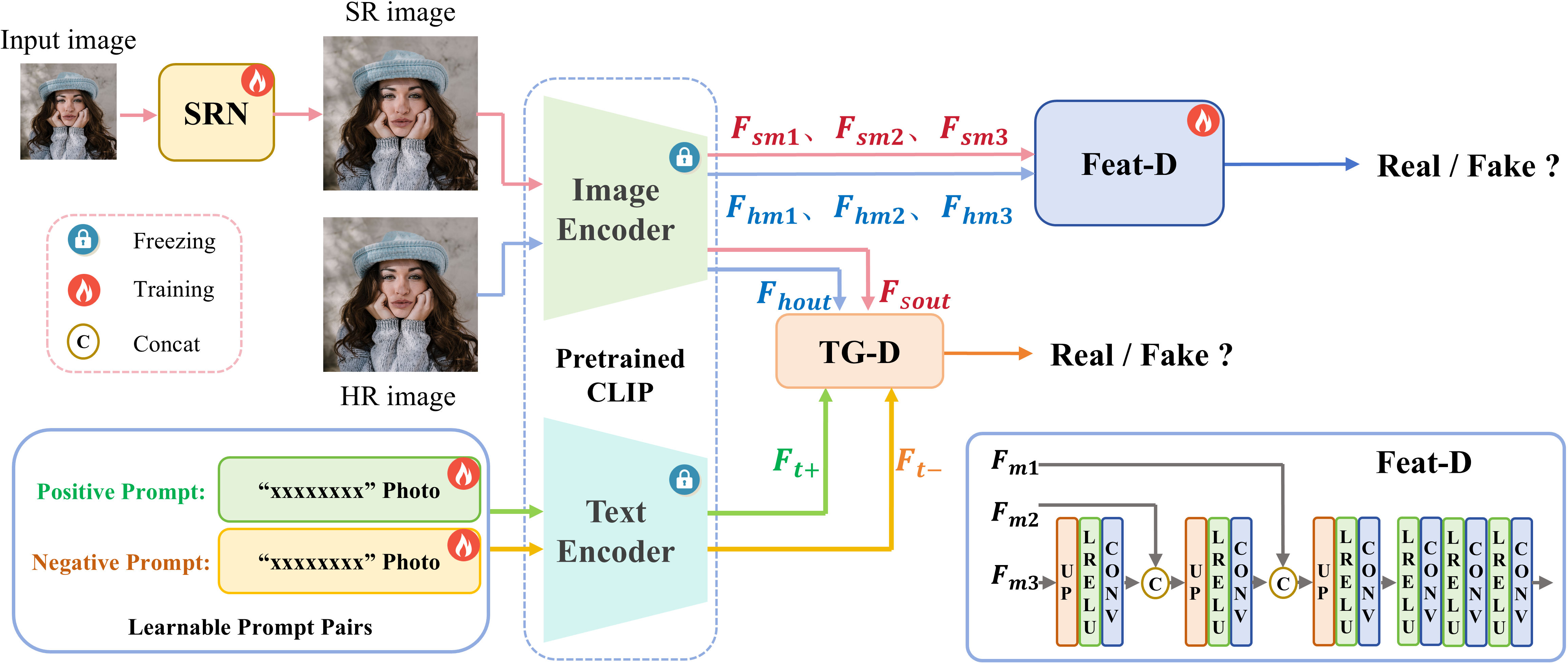}

   \caption{Framework of the proposed SFD. SFD consists of a feature discriminator (Feat-D) and a text-guided discrimination (TG-D) method, Feat-D and TG-D are used to perform discrimination on the middle features and final output features from CLIP, respectively.}
   \label{fig.clipd}
\end{figure*}

In this section, we provide a detailed explanation and analysis of our method. Firstly, we describe the overall framework of our method in \cref{Overall Framework of Proposed Method}. Next, we provide a detailed introduction and analysis of the two proposed semantic feature discrimination methods in \cref{Discrimination on Semantic Features}. Lastly, we discuss how to extend our method to OU NR-IQA in \cref{Extension to Opinion-Unaware NR-IQA}. 

\subsection{Overall Framework}
\label{Overall Framework of Proposed Method}
We propose a noval semantic feature discrimination (SFD) method for perceptual SISR, aiming to achieve better perception-distortion (PD) trade-off. Differing from vanilla GAN-based SR methods that only consider coarse-grained image distributions with poor semantic relevance, our SFD focuses on the distributions of semantic features. As illustrated in \cref{fig.clipd}, given the input low-resolution image $I_{lr}$, we can obtain its corresponding SR image $I_{sr}$ through the SRN. Then, through the pre-trained CLIP \cite{CLIP} image encoder $E_{i}$, we can obtain rich semantic features of $I_{sr}$ with different scales and characteristics, including the multi-scale pixel-wise middle features $F_{smi}$, $F_{smi}$, $F_{smi}$ $\in \mathbb{R}^{C_{i} \times H_{i} \times W_{i}}$ ($i$ = 1,2,3), and the more abstract final output feature $F_{sout}$ $\in \mathbb{R}^{1 \times L}$. To better leverage these semantic features to improve the robustness of our approach, we propose two methods to separately perform discrimination on them. 

On one hand, as for the pixel-wise middle features, we design a feature discriminator (Feat-D) to distinguish between the feature distributions of SR and HR images. On the other hand, we propose a text-guided discrimination (TG-D) method to perform discrimination on $F_{sout}$, further enhancing the overall discriminative ability. With both Feat-D and TG-D, SFD is able to facilitate the SR network (SRN) to learn more fine-grained semantic feature distributions and generate more realistic and semantic-related textures.

 
\subsection{Discrimination on Semantic Features}
\label{Discrimination on Semantic Features}
\textbf{Semantic Feature Discriminator.} As for the middle features of CLIP which contain multi-scale pixel-wise semantic information \cite{Sed}, we first design a feature discriminator, Feat-D, to distinguish between the feature distributions of SR and HR images, encouraging the semantic feature distribution of SR images to match that of HR images. Specifically, as shown in Figure \ref{fig.clipd}, we can obtain the middle features $F_{sm1}$, $F_{sm2}$, $F_{sm3}$ of SR image $I_{sr}$, and $F_{hm1}$, $F_{hm2}$, $F_{hm3}$ of HR image $I_{hr}$ through the freezing CLIP image encoder $E_{i}$. Then, the Feat-D is used to perform discrimination on those pixel-wise semantic features. During the adversarial training process, the Feat-D and the SRN will be optimized respectively using the following adversarial loss functions:
\[ 
\begin{aligned}
L_{D_{fea}}& = \mathbb{E}_{F_{hr} \sim P_{F_{hr}}}log(1 - D(F_{hm1}, F_{hm2}, F_{hm3}))\\& + \mathbb{E}_{F_{sr} \sim P_{F_{sr}}}log(D(F_{sm1}, F_{sm2}, F_{sm3})),
\label{eq1}
\end{aligned}
\tag{1}
\]
\[ 
\begin{aligned}
{L}_{G_{fea}} & = \mathbb{E}_{F_{hr} \sim P_{F_{hr}}}log(D(F_{hm1}, F_{hm2}, F_{hm3}))\\& + \mathbb{E}_{F_{sr} \sim P_{F_{sr}}}log(1-D(F_{sm1}, F_{sm2}, F_{sm3})),
\label{eq2}
\end{aligned}
\tag{2}
\]
where \(P_{F_{hr}}\) and \(P_{F_{sr}}\) represent the feature distributions of HR and SR images, respectively. Benefiting from the excellent discriminative ability of Feat-D, the SRN can learn more fine-grained semantic feature distributions aligned with HR images, generating more semantically relevant textures.

\textbf{Learnable Prompt Pairs Driven Text-Guided Discrimination.} Discriminating the final output feature $F_{sout}$ which is more global and abstract is expected to further enhance the overall performance of our method. However, differing from the middle features, $F_{sout}$ lacks the pixel-wise semantic information due to its more abstract nature, making it less suitable for Feat-D's pixel-wise discrimination process. Therefore, leveraging the inherent visual-language connection of CLIP, we propose a text-guided discrimination method, TG-D, to perform discrimination on $F_{sout}$. Specifically, inspired by~\cite{CLIPIQA, Clipscore}, we introduce learnable prompt pairs (LPP) into our method. As shown in \cref{fig.clipd}, given the positive prompt and negative prompt, we can obtain the text features $F_{t+}$ and $F_{t-}$ through the CLIP text encoder $E_{t}$. Additionally, the final output features of SR image $F_{sout}$ and HR image $F_{hout}$ can be got from $E_{i}$. Then, we adopt the same computation method as in \cite{CLIPIQA} to calculate the relative cosine similarity score between $F_{sout}$ and the text features. This takes two steps: firstly, the cosine similarities $s_{+}$ between $F_{sout}$ and $F_{t+}$, $s_{-}$ between $F_{sout}$ and $f_{t-}$ can be respectively calculated as following: 
\[ 
\begin{aligned}
{s}_{+} = \dfrac{F_{sout} \odot F_{t+}}{\left | \left |F_{sout}\right | \right | \cdot \left | \left |F_{t+}\right | \right |},\  {s}_{-} = \dfrac{F_{sout} \odot F_{t-}}{\left | \left |F_{sout}\right | \right | \cdot \left | \left |F_{t-}\right | \right |},
\label{eq3}
\end{aligned}
\tag{3}
\]
where $\odot$ denotes vector dot-product and $\left | \left | \cdot \right | \right |$ represents $L_{2}$ norm. Then, Softmax is used to calculate the final score:
\[ 
\begin{aligned}
{s_{sr}} = \dfrac{e^{s_{+}}}{e^{s_{+}}+e^{s_{-}}}.
\label{eq4}
\end{aligned}
\tag{4}
\] 
As for $F_{hout}$ of HR image, we can also adopt \cref{eq3} and \cref{eq4} to calculate the cosine similarity score $s_{hr}$. Then, we utilize $s_{sr}$ and $s_{hr}$ to train the LPP and SRN in an adversarial manner:
\[ 
\begin{aligned}
L_{D_{tg}} = \mathbb{E}_{F_{hr} \sim P_{F_{hr}}}log(1 - {s}_{hr}) + \mathbb{E}_{F_{sr} \sim P_{F_{sr}}}log({s}_{sr}),
\label{eq5}
\end{aligned}
\tag{5}
\]
\[ 
\begin{aligned}
{L}_{G_{tg}} = \mathbb{E}_{F_{hr} \sim P_{F_{hr}}}log({s}_{hr})  + \mathbb{E}_{F_{sr} \sim P_{F_{sr}}}log(1 - {s}_{sr}),
\label{eq6}
\end{aligned}
\tag{6}
\]
where \(L_{D_{tg}}\) and \(L_{G_{tg}}\) represent the adversarial loss functions for LPP and SRN, respectively. During the adversarial training process, LPP will be more and more sensitive to image quality and TG-D will be able to discriminate $F_{sout}$ in the form of assigning a global score. Finally, the SRN is optimized by the combination of the following losses as:
\[ 
\begin{aligned}
L_{total}& = \lambda_{1} \left | \left | I_{hr}, I_{sr} \right | \right |_{1} + \lambda_{2}\left | \left | \phi_{i}(I_{hr}),  \phi_{i}(I_{sr}) \right | \right |_{1} \\
 &+ \lambda_{3}{L}_{G_{fea}} + \lambda_{4}{L}_{G_{tg}},
\label{eq7}
\end{aligned}
\tag{7}
\]
where $ \left | \left | \cdot \right | \right |_{1}$ represents the $L_1$ norm, $\phi_{i}(\cdot)$ is the layers of pre-trained VGG-19 network\cite{VGG}, $\lambda_{1}$, $\lambda_{2}$, $\lambda_{3}$, $\lambda_{4}$ represent the hyperparameters for the above losses.

\begin{figure}[t]
  \centering
  \setlength{\abovecaptionskip}{0.1cm}
   \includegraphics[width=1.0\linewidth]{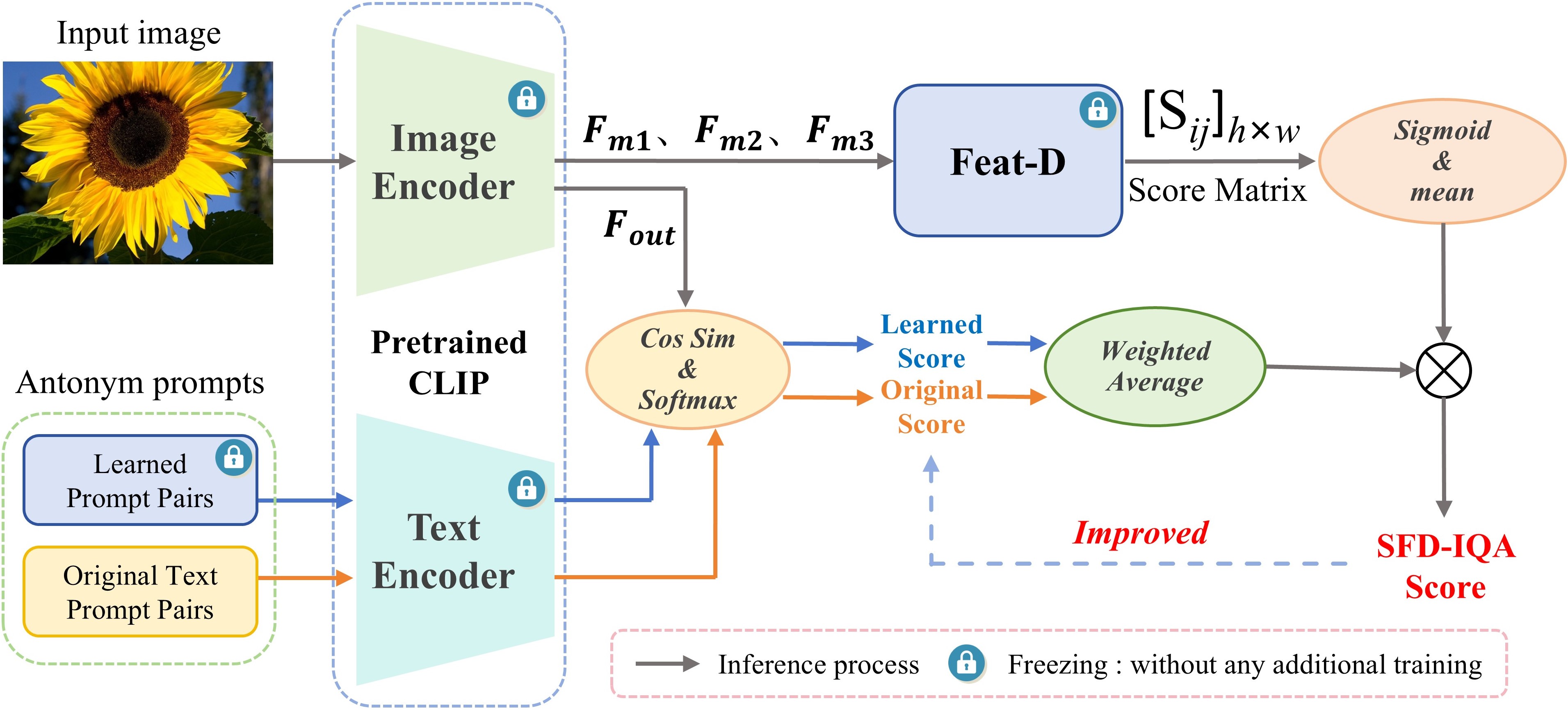}

   \caption{Framework of the proposed SFD-IQA. SFD-IQA is directly based on the well trained Feat-D and LPP, and it doesn't require any additional task-specific training on IQA datasets. }
   \label{fig.clipdiqa}
\end{figure}

\subsection{Extension to Opinion-Unaware NR-IQA}
\label{Extension to Opinion-Unaware NR-IQA}

During the process where the Feat-D and TG-D learns to distinguish the distributions of semantic features in an adversarial manner, they simultaneously and implicitly learn to assess image quality. By integrating the well trained Feat-D and LPP, we design a novel OU NR-IQA method, SFD-IQA. As shown in Fig.~\ref{fig.clipdiqa}, given an image $x$, we can obtain its middle semantic features $F_{m1}$, $F_{m2}$, $F_{m3}$, and the final output feature $F_{out}$ through the CLIP image encoder $E_{i}$. Then, Feat-D will convert $F_{m1}$, $F_{m2}$, $F_{m3}$ into a score matrix:
\[
\begin{aligned}
[{s}_{ij}]_{h \times w} = D(F_{m1}, F_{m2}, F_{m3}),
\label{eq8}
\end{aligned}
\tag{8}
\]
where $[{s}_{ij}]_{h \times w}$ denotes the score matrix. Additionally, according to Eq.~\ref{eq2} and Eq.~\ref{eq3}, we adopt $F_{out}$ and LPP to calculate the similarity score $s_{lp}$. Due to CLIP’s outstanding contrastive learning training scheme, $F_{out}$ is already sensitive to quality-related text prompts. To further enhance the robustness of SFD-IQA, we also adopt the original antonymic text prompts used in CLIP-IQA (e.g., “Good
photo” and “Bad photo”) to calculate the similarity score $s_{o}$ with $F_{out}$. Then we perform a weighted average on $s_{o}$ and $s_{lp}$:
\[ 
\begin{aligned}
{s}_{aver} = \alpha_{1}{s}_{o} + \alpha_{2}{s}_{lp},
\end{aligned}
\tag{9}
\] 
where $\alpha_{1}$ and $\alpha_{2}$ denotes the weight coefficients. Finally, following~\cite{RecycleD}, we adopt the product of ${s}_{aver}$ and the average of $[{s}_{ij}]_{h \times w}$ as the final IQA score ${s}_{d}$:
\[ 
\begin{aligned}
{s}_{d} = ({s}_{aver}) \times \dfrac{\sum_{i=0}^{h}\sum_{j=0}^{w}(sigmoid([{s}_{ij}]_{h \times w}))}{h \times w},
\end{aligned}
\tag{10}
\] 
where $sigmoid$ denotes the Sigmoid function to adjust the score range to [0, 1]. Notably, we directly apply the Feat-D and TG-D to the above IQA process without any additional task-specific training. Benefiting from the semantic-related discriminative abilities of Feat-D and TG-D, SFD-IQA can significantly improve the OU NR-IQA performance.
    
\section{Experiments}

\begin{table*}[h]
\centering
\caption{Quantitative comparison of our method vs. other SOTA methods for $\times 4$ SR task. The best and the second-best are marked in \textcolor{red}{red} and \textcolor{blue}{blue}, respectively.}
\label{table.Quantitative comparison Classical SISR}
\scalebox{0.72}{
    \centering
    \begin{tabular}{c|c|cccccccccc|cc}
    \toprule
    \multirow{2}{*}{Benchmark} & \multirow{2}{*}{Metric} & RankSR&SRGAN &SFTGAN& ESRGAN&SPSR & LDL & DualFormer & WGSR & SeD & RRDB &SwinIR& SwinIR \\
     & &GAN \cite{RankSRGAN}&\cite{SRGAN}& \cite{SFTGAN_OST} &\cite{ESRGAN} & \cite{SPSR} &\cite{LDL} & \cite{DualFormer}  & \cite{WGSR} & \cite{Sed} & +Ours & +$L_{GAN}$ & +Ours \\

    \midrule
    \multirow{4}{*}{Set5} & PSNR $\uparrow$ & 29.65 & 30.52&30.06 & 30.44&30.40 & 31.03 & 31.30 & \textcolor{blue}{31.33}  & 31.22& \textcolor{red}{31.63} &  \textcolor{blue}{31.10} & \textcolor{red}{31.78}\\
     & SSIM $\uparrow$ & 0.838 & 0.863&0.848 & 0.852&0.844 & 0.861 & \textcolor{blue}{0.869} &  0.866 & 0.868 & \textcolor{red}{0.877}& \textcolor{blue}{0.869}  &\textcolor{red}{0.877}\\
     & LPIPS $\downarrow$ & 0.070 & 0.069&0.080 & 0.074&0.069 & 0.066 & 0.066 & 0.065  & \textcolor{blue}{0.064} & \textcolor{red}{0.063} & \textcolor{blue}{0.068} &\textcolor{red}{0.059}\\
     & DISTS $\downarrow$ & 0.104 & 0.108&0.109 & 0.097&\textcolor{red}{0.093 }& \textcolor{red}{0.093} & 0.093 & 0.110 & 0.094 & 0.098 &  \textcolor{blue}{0.098}  &\textcolor{red}{0.089}\\
    \midrule
    \multirow{4}{*}{Set14} & PSNR $\uparrow$ & 26.45 & 27.01&26.74 & 26.28&26.64 & 27.10 & 27.39 & \textcolor{blue}{27.40} & 27.37 & \textcolor{red}{27.79} &  \textcolor{blue}{27.05} & \textcolor{red}{27.94}\\
     & SSIM $\uparrow$ & 0.703 & 0.728 &0.718& 0.699&0.714 & 0.735 & \textcolor{blue}{0.739} & 0.739  & 0.736& \textcolor{red}{0.752} & \textcolor{blue}{0.731} &\textcolor{red}{0.754}\\
     & LPIPS $\downarrow$ & 0.135  & 0.123 &0.131& 0.131 &0.135& 0.131 & 0.120 & 0.138  & \textcolor{red}{0.117} & \textcolor{red}{0.117} & \textcolor{blue}{0.120} & \textcolor{red}{0.118}\\
     & DISTS $\downarrow$ & 0.110 & 0.109 &0.113& 0.099&0.099 & 0.098 &\textcolor{red}{0.092} &0.112 & \textcolor{blue}{0.093} &  0.096& \textcolor{blue}{0.098} &\textcolor{red}{0.091}\\
    \midrule
    \multirow{4}{*}{Urban100} & PSNR $\uparrow$ & 24.47  & 25.02&24.34 & 24.35&24.80 & 25.49 & 25.69 & 25.78  &\textcolor{blue}{25.93} &\textcolor{red}{26.18} & \textcolor{blue}{25.83} &\textcolor{red}{26.57}\\
     & SSIM $\uparrow$ & 0.729 &  0.778 &0.724 & 0.734&0.747 & 0.767 & 0.773 &  \textcolor{blue}{0.781}  &0.780 &\textcolor{red}{0.787} & \textcolor{blue}{0.785} &\textcolor{red}{0.798}\\
     & LPIPS $\downarrow$ & 0.138  & 0.130&0.134 & 0.123 &0.121& 0.110 & 0.115 &  0.135  & \textcolor{red}{0.107} &\textcolor{blue}{0.109} &  \textcolor{blue}{0.100} & \textcolor{red}{0.099}\\
     & DISTS $\downarrow$ & 0.104  & 0.110&0.106 & 0.088&0.086 & 0.082 & 0.085 & 0.108  &\textcolor{red}{0.079} & \textcolor{blue}{0.082} & \textcolor{blue}{0.081}&  \textcolor{red}{0.079}\\
    \midrule
    \multirow{4}{*}{DIV2K100} & PSNR $\uparrow$ & 28.03 & 28.18&28.09 & 28.20&28.18 & 28.96 & 29.25& 29.19 &  \textcolor{blue}{29.27} &\textcolor{red}{29.75} &  \textcolor{blue}{28.99} &\textcolor{red}{29.90}\\
     & SSIM $\uparrow$ & 0.767  & 0.788 & 0.771&0.777 &0.772& 0.795 & 0.802 &  \textcolor{blue}{0.804}  & 0.803 & \textcolor{red}{0.815} &  \textcolor{blue}{0.803}&\textcolor{red}{0.818}\\
     & LPIPS $\downarrow$ & 0.121  & 0.122 &0.133& 0.115&0.113 & 0.101 & 0.097  & 0.104 & \textcolor{red}{0.094} &\textcolor{blue}{0.097} & \textcolor{red}{0.094} & \textcolor{blue}{0.099}\\
     & DISTS $\downarrow$ & 0.064  & 0.063&0.074 & 0.059&0.055& \textcolor{red}{0.053} & 0.056  & 0.054 & 0.054 & \textcolor{red}{0.053} & \textcolor{red}{0.050} & \textcolor{red}{0.050}\\
    \bottomrule
    \end{tabular}
    }
\end{table*}

\begin{figure*}[h]
  \centering
  \setlength{\abovecaptionskip}{0.1cm}
   \includegraphics[width=1.0\linewidth]{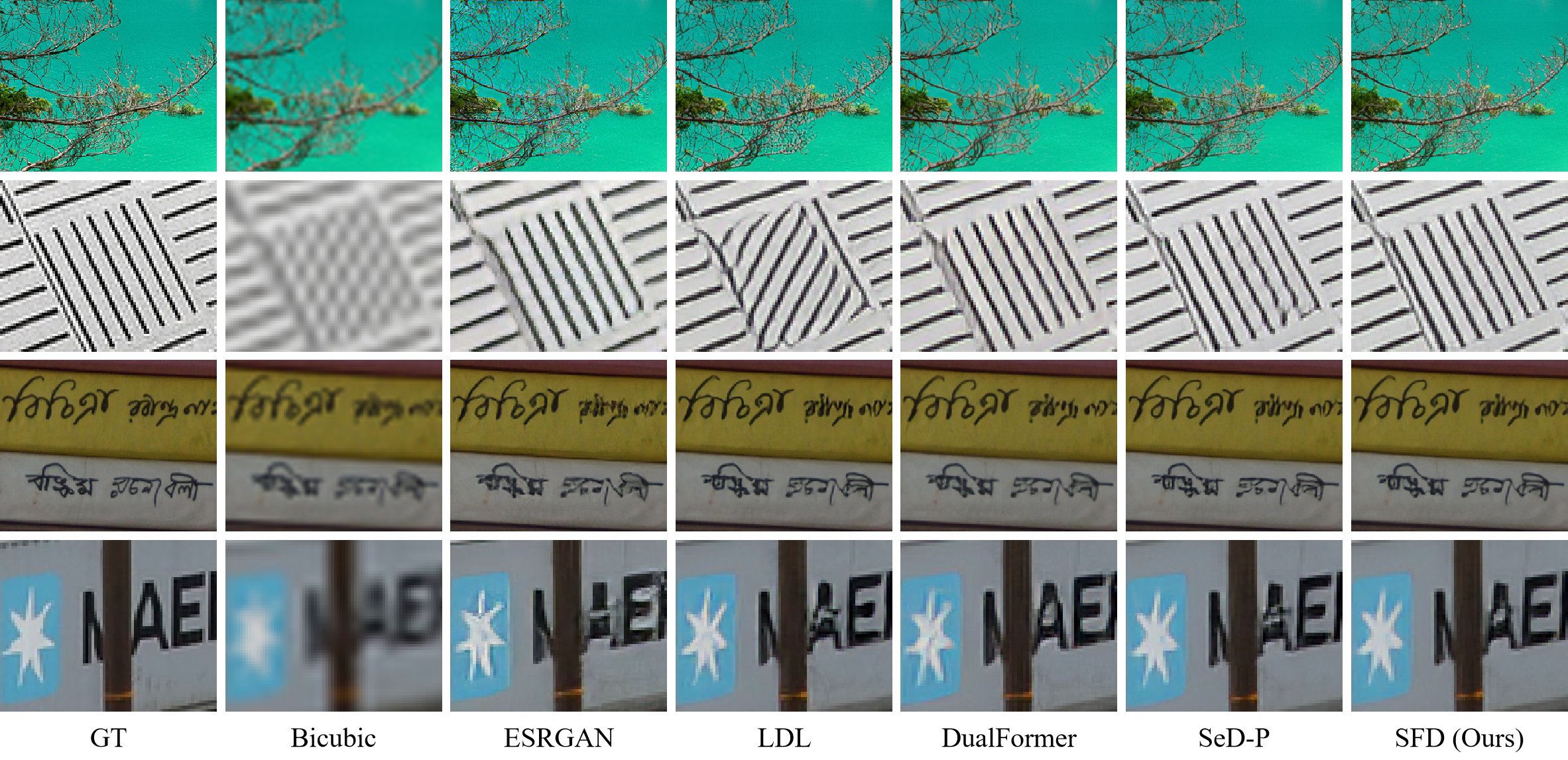}
   \caption{Visual comparison of different GAN-based SR methods on ×4 super-resolution.}
   \label{fig.classic sisr results}
\end{figure*}

\begin{figure*}[t]
  \centering
  \setlength{\abovecaptionskip}{0.1cm}
   \includegraphics[width=1.0\linewidth]{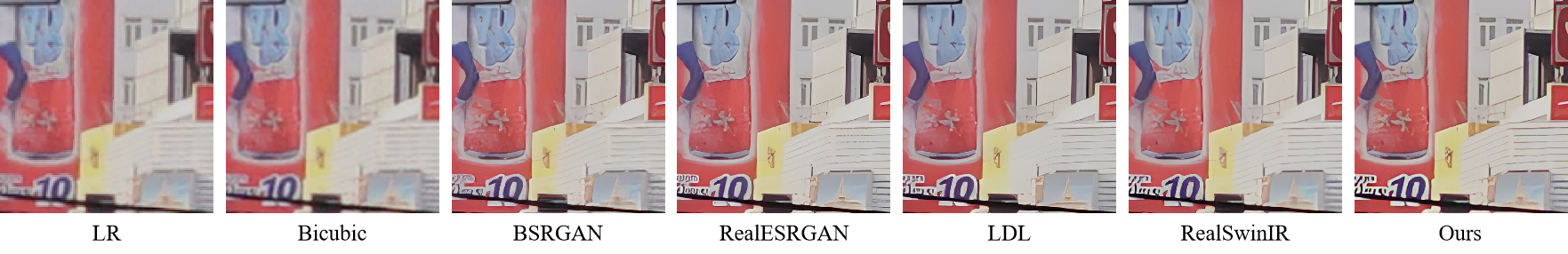}
   \caption{Visual comparison of different GAN-based real-world SR methods on ×4 super-resolution.}
   \label{fig.real-world sisr results}
\end{figure*}

\subsection{Experimental Setup}

\textbf{Training Datasets.} For classical SISR, following previous works~\cite{Sed, DualFormer, FFTGAN}, we train our network using 3450 images from DF2K~\cite{DIV2K, F2K}, and the bicubic downsampling is used to synthesize LR-HR training pairs. For real-world SISR and OU NR-IQA, following~\cite{RecycleD, DualFormer, Real-esrgan}, we adopt DF2K and OutdoorSceneTraining~\cite{SFTGAN_OST} as the training dataset, and the LR-HR training pairs are generated by the degradation pipeline of Real-ESRGAN~\cite{Real-esrgan}.

\textbf{Test Datasets and Evaluation Metrics.} For classical SISR, we test on 4 commonly used benchmarks: Set5~\cite{sET5}, Set14~\cite{SET14}, Urban100~\cite{URBAN100}, and DIV2K validation set~\cite{DIV2K}, using  4 metrics to comprehensively evaluate the performance of different methods, including fidelity metrics: PSNR and SSIM~\cite{SSIM} (calculated on the Y channel in YCbCr space), and perceptual quality metrics: LPIPS~\cite{LPIPS}, DISTS~\cite{DISTS}. For real-world SISR, following~\cite{StableSR, SeeSR}, we employ 2 center-cropped real-world datasets, RealSR~\cite{RealSR} and DRealSR~\cite{DRealSR}. We also create another dataset, RealD2K, by degrading DIV2K validation set under the same pipeline as that in training. We add another metric NIQE~\cite{NIQE} for real-world SISR. For OU NR-IQA, following~\cite{RecycleD, DualFormer}, we use two categories of IQA datasets, includinng SR IQA datasets: PIPAL~\cite{PIPAL}, Ma’s dataset~\cite{MaDATASET}, and authentically distorted IQA datasets: KonIQ-10k~\cite{KonIQ-10k}, LIVE-itW~\cite{CLIVE}. To evaluate the IQA performance, we adopt three widely used metrics: PLCC, SRCC and KRCC~\cite{KRCC}.

\textbf{Implementation Details.} For the generator, following~\cite{Sed, Real-esrgan}, we use the competitive RRDB~\cite{ESRGAN} as the backbone. We implement the experiments on NVIDIA 4090 GPUs with PyTorch~\cite{pytorch}. We initialize the parameters with the pre-trained fidelity-oriented model. We use Adam~\cite{adam} optimizer to train the network with a learning rate of $10^{-4}$.

\subsection{Results on Classical SISR}
To validate the effectiveness of our method, we compare it with several SOTA GAN-based SR methods, including \cite{SRGAN,RankSRGAN,ESRGAN,SPSR,LDL,DualFormer,WGSR,Sed}. As shown in \cref{table.Quantitative comparison Classical SISR}, compared to other approaches, our method achieves the best PSNR and SSIM scores on all benchmarks, while also holding highly competitive perceptual metrics, meaning that our method can sacrifice less fidelity in exchange for improved perceptual quality. To validate generalizability, we also employ another Transformer-based SR backbone, SwinIR \cite{SWINIR}. It's clearly that our method also demonstrates superior performance with SwinIR. \cref{fig.classic sisr results} presents the visual comparison, indicating that our method generates more realistic semantic textures with fewer artifacts than other methods.

\begin{table}[t]
\centering
\setlength{\abovecaptionskip}{0.1cm}
\captionsetup{skip=5pt}
\caption{Quantitative comparison of our method vs. other state-of-the-art real-world SISR methods for $\times 4$ SR task. The best and the second-best are marked in \textcolor{red}{red} and \textcolor{blue}{blue}, respectively.}
\label{table.Quantitative comparison Real-world SISR}
    \scalebox{0.68}{
    \centering
    \begin{tabular}{ccccccc}
    \toprule
    \multirow{2}{*}{Benchmark} & \multirow{2}{*}{Metrics}
     & BSRGAN & RealESR & LDL & RealSwinIR &  SFD \\
    & & \cite{BSRGAN} & GAN~\cite{Real-esrgan} & \cite{LDL} & \cite{SWINIR} & (Ours) \\
    \midrule
    \multirow{5}{*}{RealD2K} & PSNR $\uparrow$ & \textcolor{red}{23.51} & \textcolor{blue}{23.50} & 22.94 & 22.67 & 23.40 \\
     & SSIM $\uparrow$ & 0.585 & \textcolor{blue}{0.604} & 0.597 & 0.581 & \textcolor{red}{0.606} \\
     & LPIPS $\downarrow$ & 0.379 & \textcolor{blue}{0.368} & 0.373  & 0.368 & \textcolor{red}{0.358}  \\
     & DISTS $\downarrow$ &  0.280 & \textcolor{red}{0.278} & 0.284 & 0.279&\textcolor{red}{0.278}  \\
     & NIQE $\downarrow$ & \textcolor{red}{7.537} & 8.369 & 8.384 &\textcolor{blue}{7.866}& 8.600 \\
    \midrule
    \multirow{5}{*}{RealSR} & PSNR $\uparrow$ & \textcolor{red}{26.39} & 25.69 & 25.28 & 25.90& \textcolor{blue}{26.12} \\
     & SSIM $\uparrow$ &0.765 & 0.762 & 0.757 & \textcolor{blue}{0.768}& \textcolor{red}{0.772} \\
     & LPIPS $\downarrow$ & 0.267 & 0.273 & 0.275 & \textcolor{blue}{0.262}&\textcolor{red}{0.256}  \\
     & DISTS $\downarrow$ & 0.212 & 0.206 & 0.212 &\textcolor{red}{ 0.200}&\textcolor{red}{0.200}  \\
     & NIQE $\downarrow$ & \textcolor{red}{5.657} & 5.830 & 5.992 & 5.926&\textcolor{blue}{5.797} \\
    \midrule
  \multirow{5}{*}{DRealSR} & PSNR $\uparrow$ & \textcolor{blue}{28.75} & 28.64 & 28.21 &28.21& \textcolor{red}{29.20}  \\
     & SSIM $\uparrow$ & 0.803 & 0.805 & \textcolor{blue}{0.813}& 0.798& \textcolor{red}{0.815}\\
     & LPIPS $\downarrow$ &0.288 & 0.285 & \textcolor{red}{0.279} & 0.285& \textcolor{blue}{0.284}  \\
     & DISTS $\downarrow$ & 0.214 & \textcolor{red}{0.209} & 0.213 & \textcolor{blue}{0.210}&0.221   \\
     & NIQE $\downarrow$ & \textcolor{red}{6.519} & \textcolor{blue}{ 6.693} & 7.143 & 6.766& 6.878  \\
    \bottomrule
    \end{tabular}
    }
\end{table}

\begin{table*}
    \centering
    \captionsetup{skip=5pt}
    \caption{OU NR-IQA comparison between different methods. The best and the second-best are marked in \textcolor{red}{red} and \textcolor{blue}{blue}, respectively.}
    \scalebox{0.72}{
    \centering
    \begin{tabular}{ccccccccccccc}
    \toprule
     \multirow{2}{*}{IQA Methods} & \multicolumn{3}{c}{PIPAL~\cite{PIPAL}} & \multicolumn{3}{c}{KonIQ-10k~\cite{KonIQ-10k}} & \multicolumn{3}{c}{Ma’s dataset~\cite{MaDATASET}} & \multicolumn{3}{c}{LIVE-itW~\cite{CLIVE}}\\
    \cmidrule(lr){2-13}
      & PLCC $\uparrow$ & SRCC $\uparrow$ & KRCC $\uparrow$ & PLCC $\uparrow$ & SRCC $\uparrow$ & KRCC $\uparrow$ & PLCC $\uparrow$ & SRCC $\uparrow$ & KRCC $\uparrow$ & PLCC $\uparrow$ & SRCC $\uparrow$ & KRCC $\uparrow$  \\
    \midrule
      QAC~\cite{QAC} & 0.2566& 0.2220 &0.1503 &0.3719 &0.3397 &0.2302 &0.5071 &0.4387 &0.3055 &0.2720 &0.0457 &0.0370\\
      LPSI~\cite{LPSI}&0.2897 &0.2031 &0.1363& 0.2066 &0.2239 &0.1504 &0.5293 &0.4896 &0.3543 &0.2877 &0.0834 &0.0524 \\
      NIQE~\cite{NIQE}  &0.0439 &0.0608 &0.0401& 0.4637& 0.5291& 0.3666 & 0.6427 & 0.6261 & 0.4520 & 0.4790& 0.4493& 0.3062 \\
      IL-NIQE~\cite{IL-NIQE} &0.2748 &0.2412 &0.1630& 0.5316 &0.5057 &0.3504 &0.7155&0.6548 &0.4751&0.5039 &0.4394& 0.2985\\
      SNP-NIQE~\cite{SNP-NIQE}  &0.1820 &0.1801 &0.1214& 0.6340& 0.6285& 0.4435&0.6250 &0.5621& 0.3984 &0.5201 &0.4657 &0.3162 \\
    \midrule
      dipIQ~\cite{dipIQ}  &  0.2800& 0.2304& 0.1535& 0.4429 &0.2375 &0.1594 & 0.5167 &0.4866& 0.3607 & 0.3180 &0.1774 &0.1207 \\
      RankIQA~\cite{Rankiqa}  &\textcolor{blue}{0.3698}& \textcolor{blue}{0.3715}& \textcolor{blue}{0.2522}&  0.5028 &0.4983 &0.3448& 0.5178& 0.4913& 0.3648 &0.4528 &0.4307 &0.2945 \\
       RecycleD~\cite{RecycleD} & 0.3396 &0.3331 &0.2265 & 0.6407 &0.6364&0.4468&0.5983 &0.5555 &0.3874& 0.4806 &0.4661 &0.3194  \\
      DualFormer~\cite{DualFormer}  & 0.2896 & 0.2787 & 0.1973 & 0.6543& 0.6321& 0.4435& 0.6172 & 0.6054 & 0.4044 &0.5068 &0.4897 &0.3312\\
      ContentSep~\cite{ContentSep} & - &- &- & 0.6274 &0.6401&0.4529&- &- &-& 0.5130 &0.5060 &0.3450  \\
      CLIPIQA~\cite{CLIPIQA}  & 0.2831& 0.2762& 0.1862 & 0.4330& 0.4145 &0.2821 & 0.7175& 0.7055& 0.5063 &\textcolor{blue}{0.5853}& \textcolor{blue}{0.5647}& \textcolor{blue}{0.4010}\\
      MDFS~\cite{MDFS}  & 0.2651& 0.2273& 0.1532 & \textcolor{blue}{0.7123}& \textcolor{blue}{0.7333} &\textcolor{blue}{0.5344} & \textcolor{blue}{0.7649}& \textcolor{blue}{0.7142}& \textcolor{blue}{0.5205} &0.5364& 0.4821& 0.3274\\
      SFD-IQA (Ours) & \textcolor{red}{0.4252} & \textcolor{red}{0.4092} & \textcolor{red}{0.2816} & \textcolor{red}{0.7574}& \textcolor{red}{0.7440}& \textcolor{red}{0.5443} & \textcolor{red}{0.8012}& \textcolor{red}{0.7870}& \textcolor{red}{0.5864} &\textcolor{red}{0.7111}&\textcolor{red}{0.6984}&\textcolor{red}{0.5090} \\
    \bottomrule
    \end{tabular}
    }
\label{table: NR-IQA performance comparison}
\end{table*}

\subsection{Results on Real-World SISR}
For real-world SISR, we provide quantitative and visual comparison with representative SOTA real-world SISR models including \cite{BSRGAN,Real-esrgan,LDL,SWINIR}. The quantitative comparison is demonstrated in Table \ref{table.Quantitative comparison Real-world SISR}. The results show that our SFD achieves the best SSIM score on all datasets, while also maintaining competitive perceptual metrics, especially LPIPS and DISTS, indicating that our approach can achieve better PD trade-off. As shown in \cref{fig.real-world sisr results}, our method produces SR images with sharp edges and fine textures, accompanied by fewer undesirable artifacts. The above results suggest that our SFD can distinguish complex degraded real world images and facilitate the SR network to learn more fine-grained and semantically accurate textures. 

\subsection{Results on OU NR-IQA}
To verify the effectiveness of our SFD-IQA for OU NR-IQA tasks, we compared SFD-IQA with several representative OU NR-IQA methods, including traditional methods~\cite{QAC,LPSI,NIQE,IL-NIQE,SNP-NIQE}, deep learning-based methods~\cite{dipIQ,Rankiqa,RecycleD,DualFormer,CLIPIQA,ContentSep,MDFS}. The results are presented in Table~\ref{table: NR-IQA performance comparison}. Thanks to the excellent ability of Feat-D and TG-D to distinguish between low-quality and high-quality images, learned during the real-world SISR training process, our SFD-IQA achieves the best performance across all four datasets, validating its enhanced ability to predict the perceptual quality of images.

\subsection{Ablation Studies}

\begin{figure}[t]
\setlength{\abovecaptionskip}{0.1cm}
  \centering
   \includegraphics[width=0.85\linewidth]{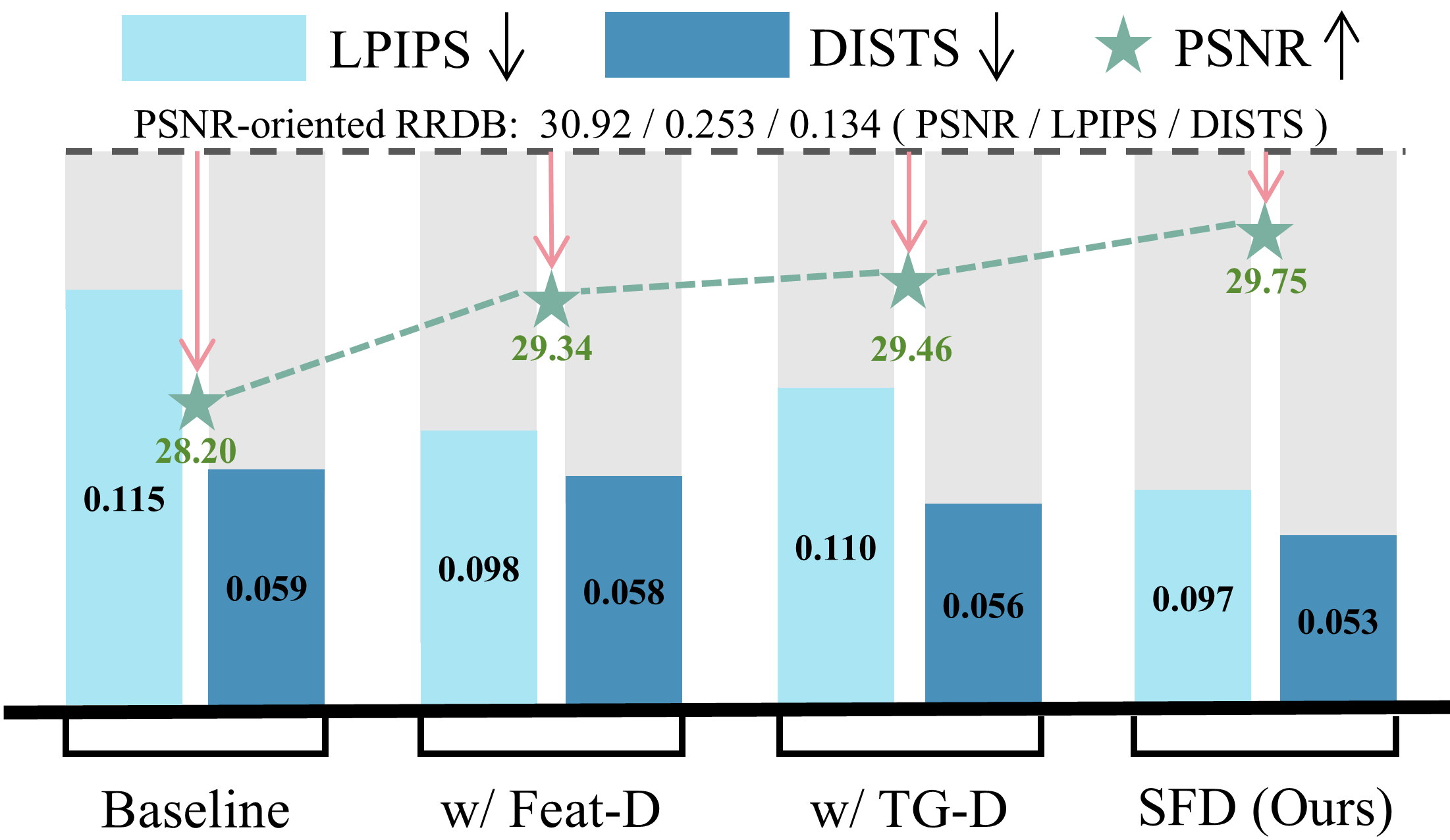}
   \caption{The effects of Feat-D and TG-D on perceptual SISR.} 
   \label{fig.ablation1}
\end{figure}

\begin{figure}[t]
  \centering
  \setlength{\abovecaptionskip}{0.2cm}
  \includegraphics[width=0.47\textwidth]{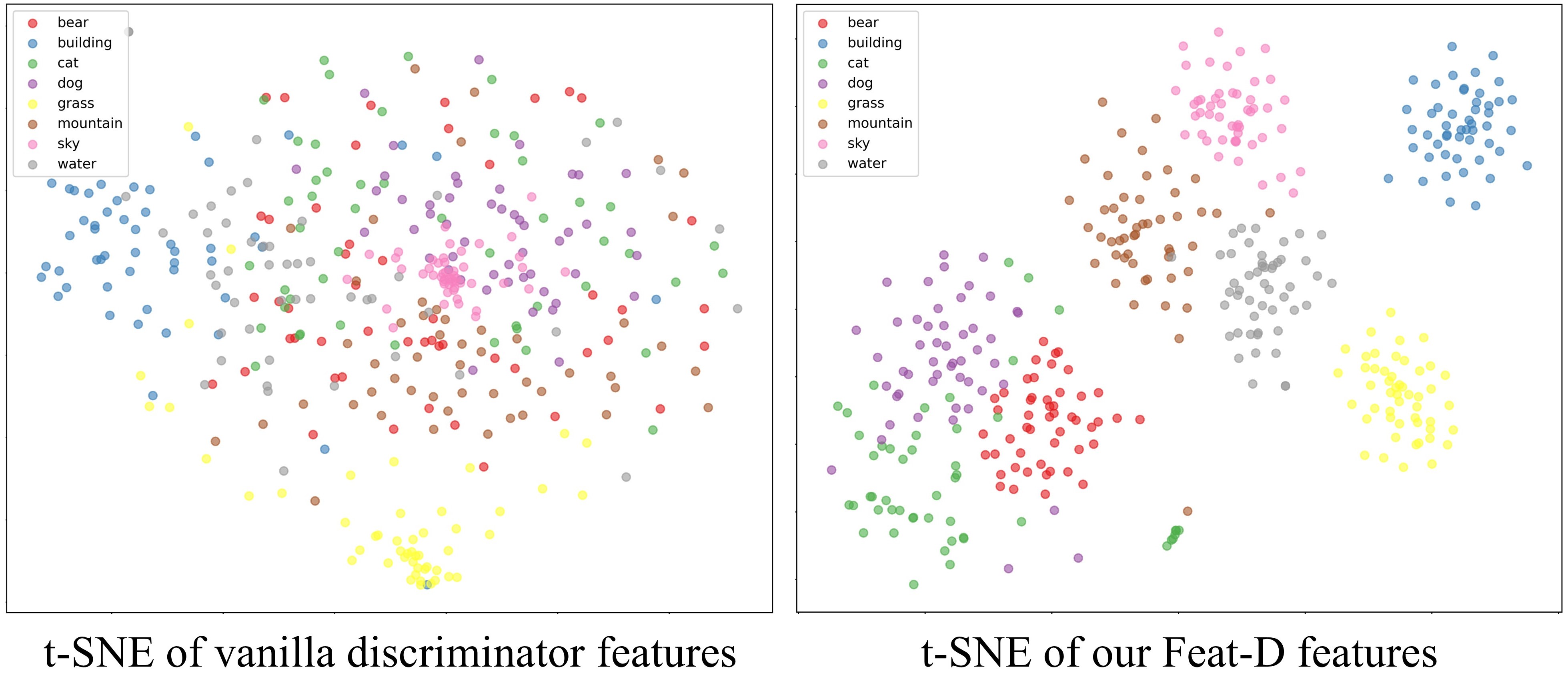}
  \caption{The t-SNE visualization of vanilla discriminator features and our Feat-D features. We select 8 categories of images from OST dataset, including “bear”, “building”, “cat”, “dog”, “grass”, “mountain”,
“sky” and “water” and 50 images for each category.}
  \label{fig.TSNE_Semantic}
\end{figure}

\textbf{The effects of Feat-D and TG-D on perceptual SISR.} 
To investigate the effectiveness of our Feat-D and TG-D, we conduct ablation experiments on them using the DIV2K100 test set. As shown in \cref{fig.ablation1}, compared to the baseline (ESRGAN), Feat-D and TG-D can both independently improve the perceptual quality with less decrease in fidelity. Furthermore, when both Feat-D and TG-D are introduced into the baseline, we can trade the minimum PSNR decrease for the best perceptual quality. The above ablation results strongly verify the superiority of Feat-D and TG-D. 

\textbf{The effects of Feat-D and TG-D in SFD-IQA.} 
To investigate the contribution of our Feat-D and TG-D to the proposed SFD-IQA, we conduct ablation studies by respectively remove Feat-D and TG-D from our SFD-IQA. 
\cref{table: Ablation on The effects of CLIP-D in CLIPDis-IQA} presents the comparative results on the 3 SR related sub-types of PIPAL~\cite{PIPAL} training set and 3 sub-types of KonIQ-10k~\cite{KonIQ-10k} dataset. Compared to the baseline without Feat-D and TG-D, it's clearly that each of Feat-D and TG-D can independently enhance the OU NR-IQA performance to a certain extent. With both Feat-D and TG-D, SFD-IQA significantly improves the performance in most cases. Especially on the GAN-based SR subset of PIPAL, it boosts the PLCC by 86.0\% and SRCC by 28.8\%. The above results demonstrate the effectiveness of Feat-D and TG-D. 

\textbf{T-SNE analysis for Feat-D.} To further demonstrate the effectiveness of our Feat-D, we use t-SNE \cite{t-SNE} to visualize the middle features of the VGG-style vanilla discriminator and our Feat-D. We select 8 categories of images from the OST \cite{SFTGAN_OST} dataset, and 50 images are randomly selected for each category. We visualize features after the 3rd upsampling layer of our Feat-D and after the 2nd BN layer of the vanilla discriminator, respectively. As shown in \cref{fig.TSNE_Semantic}, the feature clustering of our Feat-D is much better than that of the vanilla discriminator. This strongly proves that the proposed Feat-D is semantic-aware and can perform more fine-grained discrimination, encouraging the SR network to generate more more realistic semantic textures.

\begin{table}
    \centering
    \captionsetup{skip=5pt}
    \caption{OU NR-IQA performance comparison on sub-types of PIPAL training set and KonIQ-10k dataset. “Baseline" refers to the model without neither Feat-D nor TG-D. “W/ Feat-D" and “w/. TG-D" refer to the models that only include Feat-D or TG-D.}
    \scalebox{0.62}{
    \centering
    \begin{tabular}{cccccc}
    \toprule
    Metrics & Dataset. Sub-Types & Baseline & w/. Feat-D & w/. TG-D & SFD-IQA (Ours)\\
    \midrule
    \multirow{6}{*}{PLCC} & PIPAL. Trad.SR & 0.3787 & 0.3684  & \textbf{0.3910} & 0.3797 (\textcolor{red}{+0.26\%})\\
       &PIPAL. PSNR.SR&  0.4290& 0.4504 & 0.4547 & \textbf{0.4657} (\textcolor{red}{+8.55\%})\\
       &PIPAL. GAN.SR  & 0.2154 & 0.3958 &0.2430 & \textbf{0.4006} (\textcolor{red}{+86.0\%})  \\
    \cmidrule(lr){2-6}
     &KonIQ-10k. Train  & 0.7092& 0.7484& 0.7408&\textbf{0.7581} (\textcolor{red}{+6.90\%})\\
     &KonIQ-10k. Val & 0.7055 & 0.7452 &0.7350 &\textbf{0.7528} (\textcolor{red}{+6.70\%})\\
     &KonIQ-10k. Test & 0.6865 &  0.7493 & 0.7465& \textbf{0.7576} (\textcolor{red}{+10.4\%}) \\
         \midrule
    \multirow{6}{*}{SRCC} & PIPAL. Trad.SR & 0.3545 & 0.3420 & \textbf{0.3660} & 0.3522 (-0.64\%)\\
       &PIPAL. PSNR.SR&  0.4432& 0.4797 &0.4672 & \textbf{0.4932} (\textcolor{red}{+11.3\%})\\
       &PIPAL. GAN.SR  & 0.2376 & 0.3048 &0.2598 & \textbf{0.3061} (\textcolor{red}{+28.8\%})  \\
    \cmidrule(lr){2-6}
     &KonIQ-10k. Train  & 0.6849& 0.7325&0.7130&\textbf{0.7441} (\textcolor{red}{+4.92\%})\\
     &KonIQ-10k. Val & 0.6803 & 0.7264 & 0.7072&\textbf{0.7373} (\textcolor{red}{+4.51\%})\\
    & KonIQ-10k. Test & 0.6862 & 0.7359 &0.7131 & \textbf{0.7475} (\textcolor{red}{+8.89\%}) \\
    \bottomrule
    \end{tabular}
    }
    \vspace{-0.3cm}
    \label{table: Ablation on The effects of CLIP-D in CLIPDis-IQA}
\end{table}

\textbf{The effectiveness of our method on other image restoration tasks.} We preliminarily explore the effectiveness of our approach in addressing other image restoration tasks, such as image deraining and image deblurring. We select a commonly used image restoration baselines NAFNet~\cite{NAFNet}, and conduct training and testing on the Rain100L~\cite{Rain100L} and GoPro~\cite{GoPro} datasets for the two tasks. As shown in \cref{table: The effectiveness of our method on other image restoration tasks.}, compared to fidelity-oriented baseline methods and vanilla discriminator based methods, our method significantly improves perceptual metrics while maintaining a high level of fidelity on both two tasks, demonstrating the potential of our method to be integrated into other image restoration tasks for better perceptual quality. 

\begin{table}[t]
    \centering
    \captionsetup{skip=5pt}
    \caption{The effectiveness of our method on image deraining and image deblurring. “VD" means vanilla discriminator.}
    \scalebox{0.7}{
    \centering
    \begin{tabular}{ccccccc}
    \toprule
    \multirow{2}{*}{Method} & \multicolumn{3}{c}{Rain100L~\cite{Rain100L}} & \multicolumn{3}{c}{GoPro~\cite{GoPro}} \\
    \cmidrule(lr){2-7}
     & PSNR $\uparrow$ & LPIPS $\downarrow$ & DISTS $\downarrow$ & PSNR $\uparrow$ & LPIPS $\downarrow$ & DISTS $\downarrow$  \\
    \midrule
    NAFNet &\textbf{37.00} &0.022 &0.031 & \textbf{32.87} & 0.093 & 0.072\\
     + VD& 36.39& 0.015& 0.020& 32.01 & 0.068 & 0.036\\
     + Ours&36.49& \textbf{0.014}& \textbf{0.019}& 32.19 & \textbf{0.064} & \textbf{0.034}\\
    \bottomrule
    \end{tabular}
    }
    \label{table: The effectiveness of our method on other image restoration tasks.}
    \vspace{-0.15cm}
\end{table}

\section{Conclusion}
In this paper, we propose a semantic feature discrimination method, SFD, for perceptual SISR and OU NR-IQA. We first design a feature discriminator (Feat-D) to discriminate the semantic features from CLIP, effectively encouraging the feature distributions of SR images to match that of HR images. Then we propose a text-guided discrimination (TG-D) method by training the learnable prompt pairs (LPP) in an adversarial manner to further enhance the discriminative ability. Our SFD facilitates the SR network to generate more fine-grained and more realistic semantic textures, achieving better perception-distortion trade-off. Furthermore, by effectively integrating the well trained Feat-D and LPP, we significantly improve the OU NR-IQA performance without any additional task-specific training on IQA datasets. Extensive experiments on classical SISR, real-world SISR, and OU NR-IQA tasks validate the effectiveness of our approach.
\\
\\
\noindent\textbf{Acknowledgement.} This work was supported by the National Natural Science Foundation of China under Grant 62171304 and partly by the Natural Science Foundation of Sichuan Province under Grant 2024NSFSC1423, Cooperation Science and Technology Project of Sichuan University and Dazhou City under Grant 2022CDDZ-09, the TCL Science and Technology Innovation Fund under grant 25JZH008, and the Young Faculty Technology Innovation Capacity Enhancement Program of Sichuan University under Grant 2024SCUQJTX025.
\newpage

{
    \small
    \bibliographystyle{ieeenat_fullname}
    \bibliography{main}
}

\end{document}